\documentclass[conference]{IEEEtran}
\IEEEoverridecommandlockouts
\usepackage{cite}
\usepackage{amsmath,amssymb,amsfonts}
\usepackage{algorithmic}
\usepackage{graphicx}
\usepackage{textcomp}
\usepackage{xcolor}
\usepackage{subcaption}
\def\BibTeX{{\rm B\kern-.05em{\sc i\kern-.025em b}\kern-.08em
    T\kern-.1667em\lower.7ex\hbox{E}\kern-.125emX}}
\begin{document}

\title{ResAF-Net: An Anchor-Free Attention-Based Network for Tree Detection and Agricultural Mapping in Palestine}

\author{\IEEEauthorblockN{1\textsuperscript{st} Rabee Al-Qasem}
\IEEEauthorblockA{\textit{AI-Developer} \\
\textit{Ministry of Labor - GGateway.ps}\\
Nablus, Palestine \\
ORCID: 0009-0000-2608-6648}

}

\maketitle

\begin{abstract} Reliable agricultural data is essential for food security, land-use planning, and economic resilience, yet in Palestine, such data remains difficult to collect at scale because of fragmented landscapes, limited field access, and restrictions on aerial monitoring. This paper presents ResAF-Net, a satellite-based tree detection framework designed for large-scale agricultural monitoring in resource-constrained settings. The proposed architecture combines a ResNet-50 encoder, Atrous Spatial Pyramid Pooling (ASPP), a feature-fusion stage, a multi-head self-attention refinement module, and an anchor-free FCOS detection head to improve tree localization in dense and heterogeneous scenes. Trained on the MillionTrees benchmark, the model achieved 82\% Recall, 63.03\% mAP@0.50, and 35.47\% mAP@0.50:0.95 on the validation split, indicating strong sensitivity to tree presence while maintaining competitive localization quality. Beyond benchmark evaluation, we implemented the model within a web-based GIS application integrated with Palestinian cadastral data from GeoMolg, enabling tree analysis at scene, parcel, and community levels. This deployment demonstrates the practical feasibility of AI-assisted agricultural inventorying in Palestine. It provides a foundation for data-driven monitoring, reporting, and future species-level analysis of Mediterranean tree crops. \end{abstract}


\section{Introduction}

Artificial Intelligence (AI) has significantly influenced many sectors of daily life, including healthcare, law, business, and finance, which have increasingly integrated AI functionality into their workflows. On the other hand, the agricultural sector has received relatively less attention in this area, particularly among small and medium-scale farmers, despite many studies showing that the integration of AI can enhance productivity and improve monitoring, and enhance the livelihoods, food security, and economic resilience  
in several areas. This yields the importance of monitoring every aspect of this sector.
\cite{alegbeleye2025urban,wang2025harnessing}.

Among the many components of agricultural and ecological systems that could benefit from improved monitoring using AI  are trees and forests. They are particularly important for sustaining environmental and economic stability and for providing essential services, such as carbon storage, soil conservation, and regulation of the water cycle. Recent reports estimate that the global forest area in 2025 is 4.14 billion hectares, representing approximately 32\% of total land area and nearly 80\% of terrestrial ecosystems \cite{fao2025fra}. Due to the large scale of these regions, traditional manual monitoring is no longer sufficient, which has created the need for an AI-driven solution for sustainable management.

In Palestine, agriculture remains one of the main sectors on which many people depend for both income and food security. However, this sector has been severely affected by limited resources, climate change, and  political pressures \cite{abu2025agroecology}. Available data indicate that the average agricultural landholding in the Palestinian Territory declined from 18.6 dunums in 2004 to 8.6 dunums in 2020, demonstrating a significant reduction in the scale of individual agricultural operations. Rapid urban expansion has also placed increasing pressure on agricultural land, contributing to the conversion of fertile areas into built environments and further accelerating land fragmentation. This decline suggests increased land fragmentation, which may lead to reduced agricultural productivity and competitiveness. Furthermore, data show that only 6.2\% of workers were employed in agriculture as of 2023 \cite{al2024palestinian,rafeedie2026impact}, underscoring the sector's diminishing role in the labor market and raising concerns about rural livelihoods and long-term food security. At the same time, traditional data collection methods require substantial human effort and often fail to provide timely, comprehensive insights, thereby limiting the effectiveness of policies and support initiatives aimed at strengthening the sector.  As a result, the need for accurate and large-scale agricultural data has become increasingly urgent.

Building on this, AI offers a practical path toward scalable agricultural monitoring. In particular, tree detection should not be viewed merely as a monitoring task, but as a strategic data tool that can support decision-makers in designing smarter agricultural policies based on large-scale satellite-derived evidence. This is especially relevant in the MENA region, where detecting agricultural assets and crop patterns remains difficult and where accurate, up-to-date agricultural data is often lacking.

Recent advances in computer vision and remote sensing have significantly improved the performance of object detection models on aerial and satellite imagery. These developments have made automated tree detection increasingly feasible through multiple platforms, including satellite imagery, aerial photography, and uncrewed aerial vehicles (UAVs). In the Palestinian context, however, satellite-based detection serves as a critical technological workaround to the severe political and regulatory restrictions on UAV usage. While drone deployment is often prohibited or heavily constrained, satellite imagery provides one of the few viable paths for consistent, large-scale monitoring of agricultural assets.

Nevertheless, detecting trees from satellite imagery remains a challenging computer vision task, particularly in Palestine, where agricultural land is highly fragmented, tree crowns are often small relative to image resolution, and local annotated datasets are limited. In addition, many existing remote sensing approaches are developed under different environmental and agricultural conditions, which may reduce their suitability for Palestinian landscapes. In this context, developing accurate and scalable tree-detection models is not only a technical research problem but also a practical step toward building data-driven tools that support agricultural monitoring, planning, and policy design in resource-constrained environments.

To the best of our knowledge, this work represents the first model to test the applicability of anchor-free object detection methods to tree crown identification and the first system to integrate deep-learning-based agricultural monitoring with official Palestinian cadastral data (GeoMolg) for parcel-level analysis.

\section{Related Works}

Earlier methods for tree detection using AI typically relied on image processing and feature extraction, focusing primarily on color. While these methods were effective for simple images in which trees were isolated from the ground, they showed limited accuracy when applied to satellite imagery of dense forests \cite{wulder2004comparison,brandt2026chmv2}. However, with the rise of AI and computer vision, applications have become increasingly effective at detecting small objects. These models have become the go-to solutions for tree crown detection, prompting increased attention to the creation of specialized datasets to support tree detection from satellite imagery \cite{topgul2025vhrtrees}. \cite{que2026fm} developed a custom framework for tree crown segmentation in UAV images, built on top of Yolov10. Their model demonstrates strong performance and generalization across diverse public datasets. Another study \cite{jarahizadeh2026tree} developed Tree-Net, a novel deep learning model for individual tree detection. Their model outperforms existing methods by up to 41\% improved accuracy and precision. They tested their model on 25000 samples from diverse datasets, and it showed faster performance than the traditional YOLO on both training and testing. Finally, another approach compares well-known methods for tree detection, including YOLO, RetinaNet, R-CNN, and Mask R-CNN, and they developed a state-of-the-art urban tree crown detector based on a dataset in the USA. Their model shows a great performance compared to other models \cite{alegbeleye2025urban}. Traditional object detection frameworks often rely on "anchor boxes"—predefined bounding box templates of fixed scales and aspect ratios \cite{zand2022objectbox,zamboni2021benchmarking}. However, these methods frequently struggle with the dense, irregular, and overlapping tree crowns typical of Mediterranean landscapes, as they require manual tuning of anchors that may not align with varying canopy shapes. In contrast, our anchor-free approach treats detection as a per-pixel prediction task, eliminating rigid priors and allowing the model to more flexibly identify closely spaced trees in fragmented agricultural plots.

\section{Methodology}

\subsection{Dataset}

To develop a robust Tree detection model, a large and diverse dataset is required. The MillionTrees benchmark \cite{weinstein2025milliontrees,brandt2026chmv2} addresses this need by providing a multi-source dataset comprising several types of images—from high-quality satellite images where trees are easily identified, to low-quality zoomed-out images that make tree detection challenging. The dataset consists of 12 sets of human-annotated bounding boxes and two weakly supervised data sources. These sources span multiple regions, including areas in North and South America. Each data source is structured with defined training and testing splits, which we used to train the model. Detailed dataset structure will be described in section \ref{sec:training_sec}.

We decided to retain 10 sources for model training and evaluation. We excluded the other two. The weakly supervised dataset is noisy, with misclassification and higher label noise. As a result, we obtained a dataset comprising over 10,000 images and about 700,000 human-annotated bounding boxes, as shown in Table \ref{tab:treeboxes}.

\begin{table}[htbp]
\caption{Summary of Human-Verified Data Sources in MillionTrees project}
\begin{center}
\addtolength{\tabcolsep}{-2pt} 
\begin{tabular}{|l|c|c|c|}
\hline
\textbf{Source} & \textbf{Annotations} & \textbf{Images} & \textbf{Citation} \\
\hline
OAM-TCD & 266,510 & 3,854 & \cite{veitch2024oam} \\
\hline
SelvaBox & 131,438 & 1,258 & \cite{baudchon2025selvabox} \\
\hline
Radogoshi et al. (2021) & 115,683 & 1,533 & \cite{radogoshi2021tree} \\
\hline
Weecology / Univ. Florida & 83,281 & 2,084 & \cite{weinstein2021benchmark} \\
\hline
Puliti \& Astrup (2022) & 61,897 & 618 & * \\
\hline
Sun et al. (2022) & 31,228 & 148 & \cite{sun2022counting} \\
\hline
Reiersen et al. (2022) & 7,072 & 99 & * \\
\hline
Dumortier et al. (2025) & 6,816 & 490 & \cite{dumortier2025annotated} \\
\hline
NEON Benchmark & 6,625 & 186 & \cite{weinstein2019individual} \\
\hline
NEON MultiTemporal & 2,602 & 152 & \cite{weinstein2021neon} \\
\hline
\end{tabular}
\label{tab:treeboxes}
\end{center}
\end{table}

\subsection{Data Analysis and Pre-processing}

Since we are working with different data sources, the three things we need to consider first are image quality, image size, and the bounding box aspect ratio. Each presents unique challenges, especially as we integrate diverse datasets. 

\begin{figure}[htbp]
\centering
\includegraphics[width=\columnwidth]{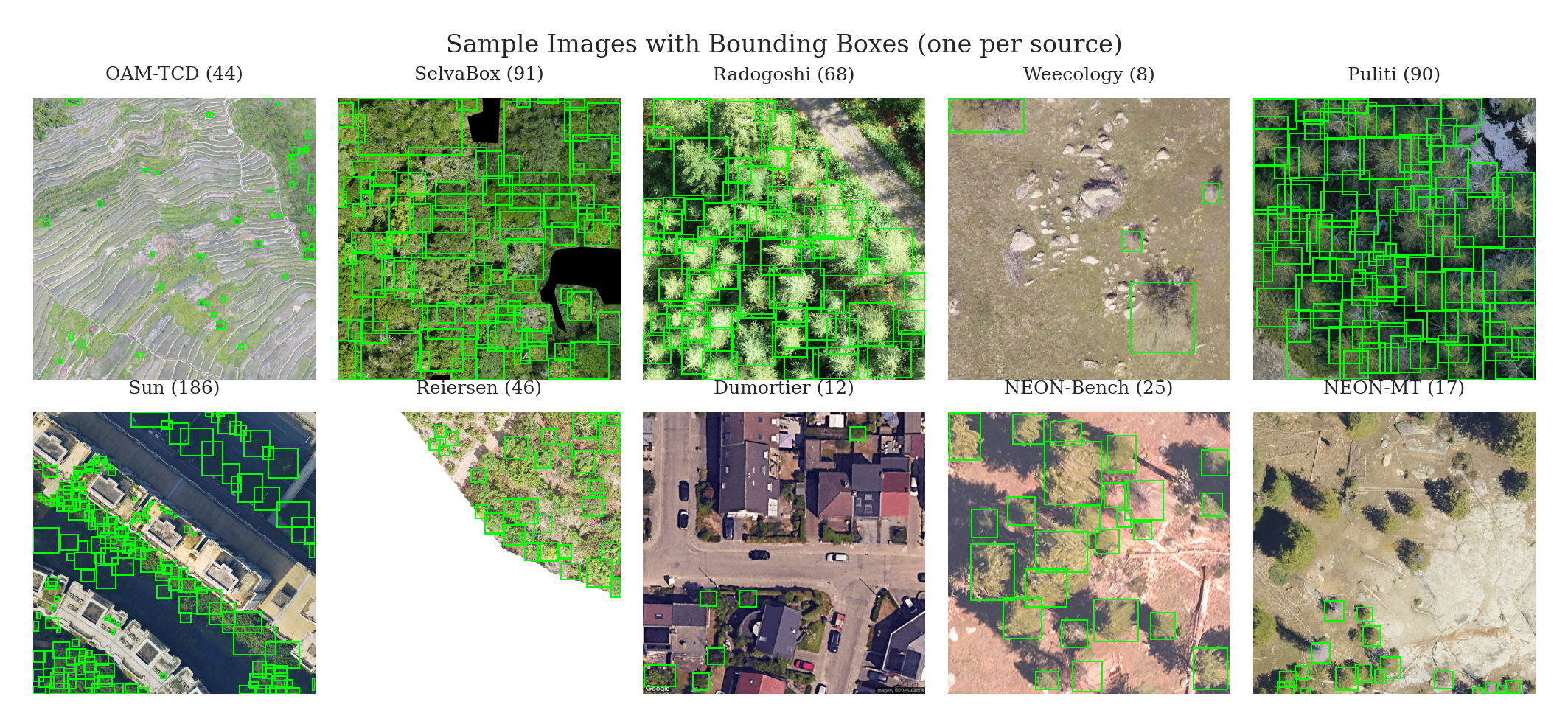}
\caption{Sample Image with its annotations from different datasets}
\label{fig:sample}
\end{figure}

\begin{figure}[htbp]
\centering
\includegraphics[width=\columnwidth, height=8cm, keepaspectratio]{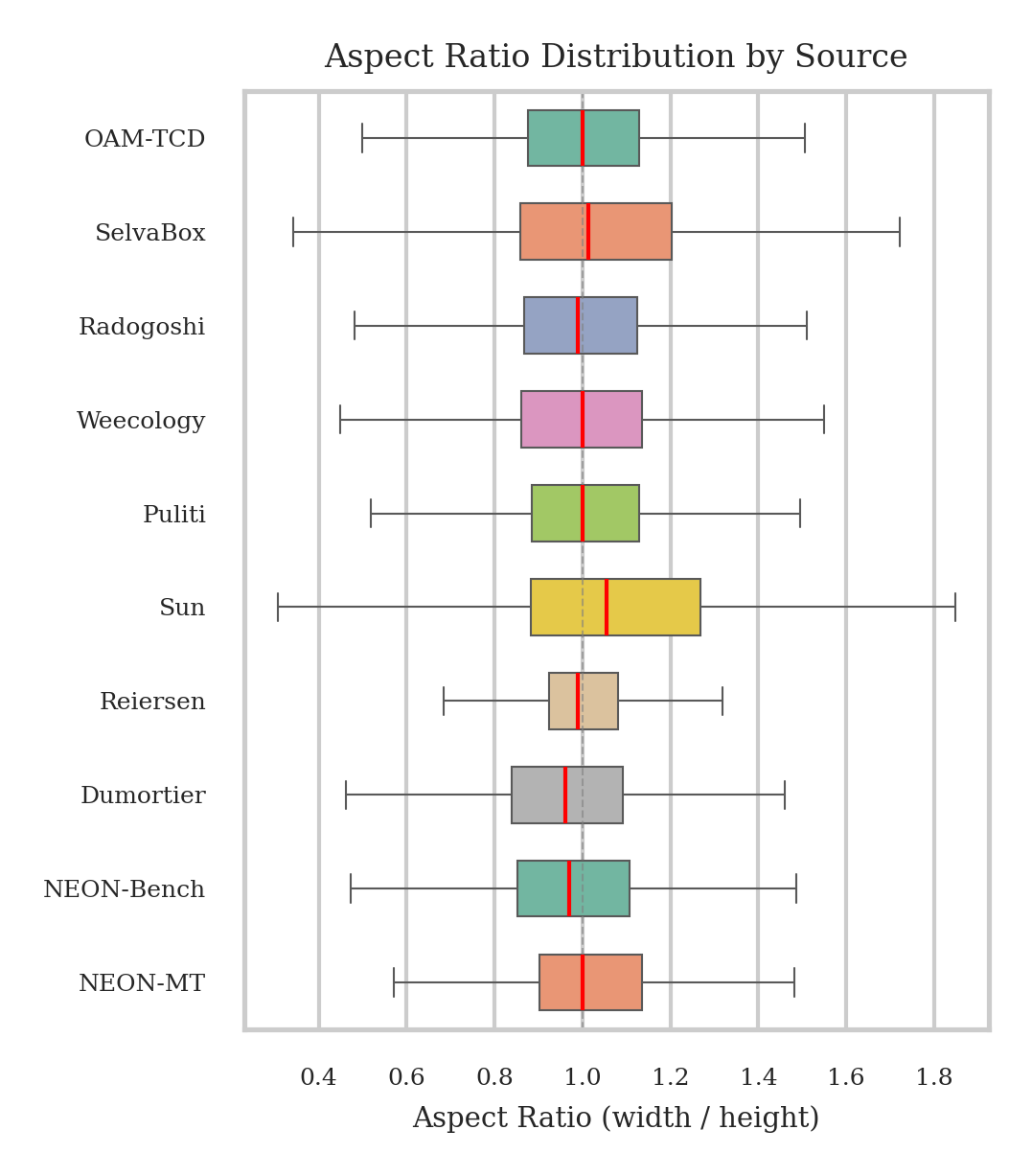}
\caption{The distribution of aspect ratio between different datasets}

\label{fig:aspect_ratio}
\end{figure}

Regarding image quality, we notice that some datasets are higher quality than others, as shown in Fig. \ref{fig:sample}, which illustrates a good variety of images, but this may pose problems for the backbones. Because satellite images can vary in lighting conditions and sharpness, data augmentation is necessary during training. Techniques such as color jitter and blur expose the model to different image qualities and reduce its sensitivity to visual inconsistencies. In addition, geometric augmentations such as rotation and flipping help the model learn tree patterns from different orientations, which can improve generalization when image details are weak or partially unclear.
Regarding image size, some datasets, such as \cite{sun2022counting}, exhibit a wide range of aspect ratios as illustrated in Fig. \ref{fig:aspect_ratio}. Image sizes vary from as large as (4000 * 4000) pixels to as small as (250 * 250) pixels. Some images are horizontal, while others are vertical. To manage this variation, we set the image input size to (640*640). Instead of resizing large images, which risks losing information, we employ a patching technique that splits large images into overlapping (640640) squares. For images smaller than 640 in either dimension, we apply padding.

\begin{figure}[htbp]
\centering
\includegraphics[width=\columnwidth,, height=8cm, keepaspectratio]{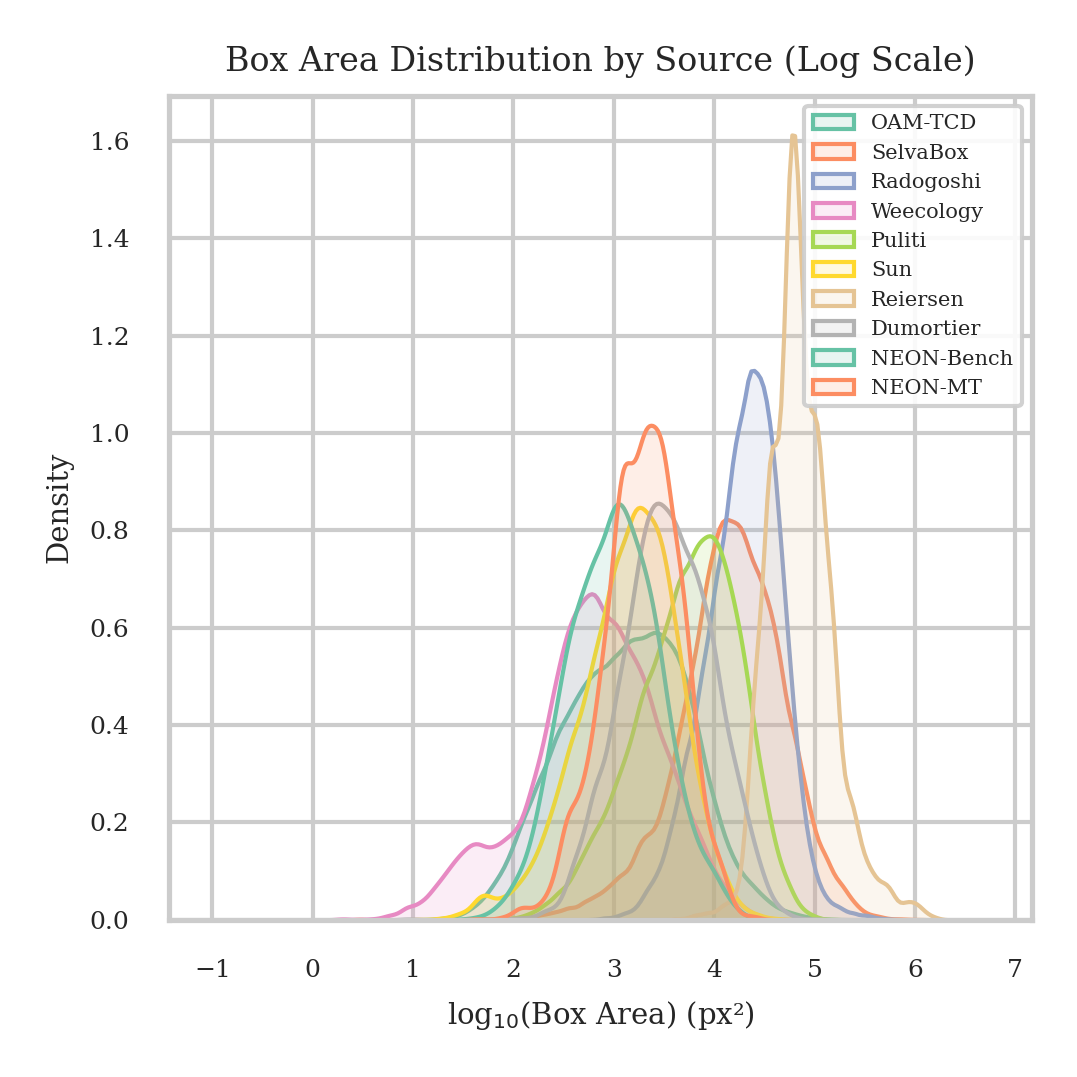}
\caption{Box Area distribution across datasets}
\label{fig:bbox_dist}
\end{figure}

Finally, regarding one of the advantages, we have different datasets in detection models having a variety of bounding box sizes, which helps the model to generalize around different tree sizes (small, medium, large sizes), and the dataset is very representative in this regard, as we can see in Fig. \ref{fig:bbox_dist}

\subsection{Model Architecture}

\begin{figure}[htbp]
\centering
\includegraphics[width=\columnwidth]{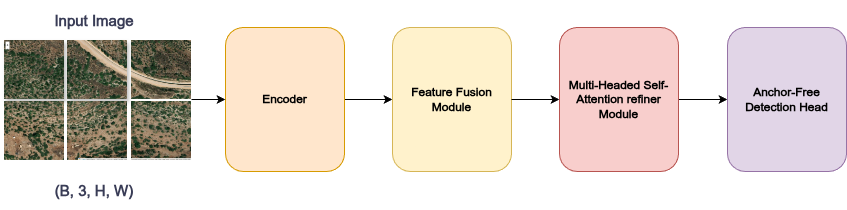}
\caption{Tree detection architecture consisting of an encoder, feature fusion module, attention refiner, and anchor-free detection head.}
\label{fig:full}
\end{figure}

To address the specific challenges of tree detection in satellite imagery, we designed a custom detection pipeline tailored to our task rather than relying entirely on a standard off-the-shelf architecture. This decision was motivated by several factors. First, our application involves characteristics that are not fully addressed by generic detection models, particularly small object sizes, background complexity, and the visual variability of trees in aerial and satellite imagery. Second, we aimed to develop a lightweight model that operates efficiently on limited computational resources, making it more suitable for inference on small GPUs and for practical deployment. Third, the model was designed to be more tolerant of the common challenges found in satellite imagery, such as variations in image quality, resolution, illumination, and scene clutter. As illustrated in Fig. \ref{fig:full}, the proposed tree detection model comprises four main modules: an encoder for hierarchical feature extraction, a feature fusion module for combining multi-scale representations, a multi-headed self-attention refiner module for enhancing salient tree-related features, and an anchor-free detection head for final bounding-box prediction.

\subsubsection{Encoder}

\begin{figure}[htbp]
\centering
\includegraphics[width=\columnwidth]{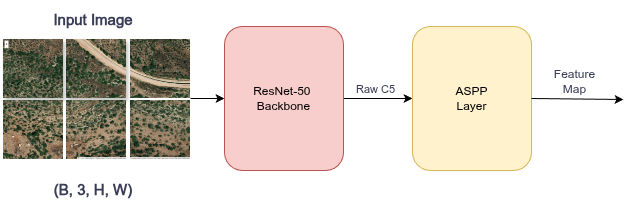}
\caption{Architecture of the proposed encoder. An ASPP module is integrated following the ResNet-50 C5 stage to extract multi-scale contextual features.}
\label{fig:encoder}
\end{figure}

The encoder plays a central role in our model because satellite imagery often contains complex land-cover patterns, varying color distributions, and, in some cases, low-quality visual details, making feature extraction challenging. Although transformer-based backbones such as Vision Transformer (ViT) and Swin Transformer have shown strong performance in visual recognition tasks, they generally require greater computational resources, making them less practical under limited GPU memory constraints \cite{dosovitskiy2020image, liu2022swin}. Therefore, we adopt a pretrained ResNet-50 backbone, which provides an efficient and well-established convolutional architecture for hierarchical feature extraction \cite{he2016deep}. In ResNet-based encoders, deeper stages progressively increase channel capacity while reducing spatial resolution, yielding more semantically rich but less spatially detailed representations. To mitigate the loss of fine-grained information at the deepest stage, we enhance the final C5 representation using an Atrous Spatial Pyramid Pooling (ASPP) module inspired by DeepLab \cite{chen2017deeplab,chen2017rethinking}, as illustrated in Fig. \ref{fig:encoder}. ASPP aggregates multi-scale contextual information via parallel atrous convolutions and image-level pooling, yielding a richer deep feature representation for objects at different scales. In addition, our design was motivated by recent studies that combined ResNet-based encoders with ASPP modules in remote sensing and related vision tasks, prompting us to replicate and adapt this integration in our architecture \cite{cai2024res50, deng2025residual}.

\subsubsection{Feature Fusion Module}

The encoder produces feature maps with different channel dimensions, and to unify these representations, we add a small Feature Pyramid Network (FPN) that transforms all feature maps into a single unified dimension, so they can then be passed to the attention layers. We also found that smoothing these feature maps significantly improves the detection of small and medium trees.

\begin{figure}[htbp]
\centering
\includegraphics[width=\columnwidth, height=5cm, keepaspectratio]{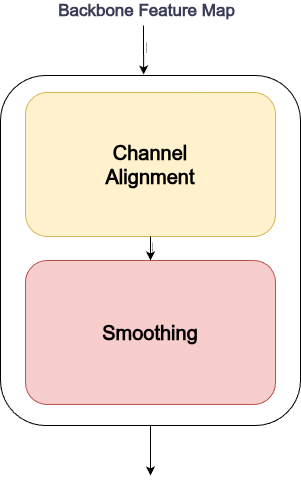}
\caption{High-level overview of the Feature Pyramid Network (FPN) neck. The encoder outputs feature maps with varying channel dimensions, which are unified via lateral convolutions and refined by smoothing operations before being passed to the attention layers.}

\label{fig:FPN}
\end{figure}

\subsubsection{Multi-Headed Self-Attention refiner Module}

In the final stage before the detection head, the feature maps produced by the feature fusion module (FPN) are passed to an attention refiner. This idea has previously been explored in Conformer-based models \cite{kiessling2024does,gulati2020conformer}. The main purpose of this layer, which is built on multi-head self-attention, is to identify the most salient signals in the image that represent trees. It achieves this by dynamically reweighting features during training, effectively amplifying relevant information while suppressing background noise. To provide the model with spatial awareness, we also incorporate a 2D positional encoding. Unlike the fixed sinusoidal encoding used in the original Transformer, our implementation employs learnable row and column embeddings, offering greater flexibility for the varying spatial dimensions of the feature pyramid.

\begin{figure}[htbp]
\centering
\includegraphics[width=\columnwidth, height=5cm, keepaspectratio]{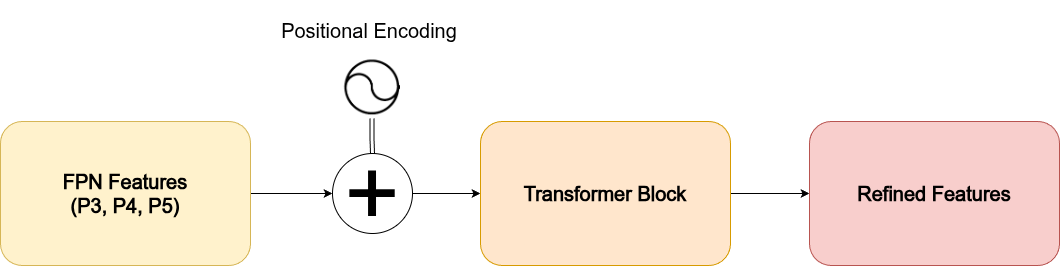}
\caption{High-level architecture of the self-attention refiner. FPN feature maps are enriched with 2D positional encoding and processed by a transformer block to generate refined features.}

\label{fig:attention}
\end{figure}

\subsubsection{Anchor-Free Detection Head}To translate the refined multi-scale features into final tree detections, we employ an anchor-free detection head based on the Fully Convolutional One-Stage (FCOS) framework \cite{tian2019fcos}. Unlike traditional detectors that rely on predefined anchor boxes, this head treats each location on the feature map as a training sample, predicting three distinct components per pixel: (1) Classification Scores, which estimate the probability of a pixel containing a tree center; (2) Regression Offsets, which calculate the four-dimensional distances $(l, t, r, b)$ from the current location to the bounding box boundaries; and (3) Centerness, a unique branch designed to predict how close the current pixel is to the center of the box it just predicted. By multiplying the classification score by the centerness value, the model effectively suppresses low-quality bounding boxes generated from locations far from the object's center. This pixel-level approach is particularly well-suited to high-resolution satellite imagery, as it eliminates the need for manual anchor tuning and provides a more flexible response to the varying scales and dense spatial distributions characteristic of forest and orchard environments.

\begin{figure}[htbp]
\centering
\includegraphics[width=\columnwidth, height=5cm, keepaspectratio]{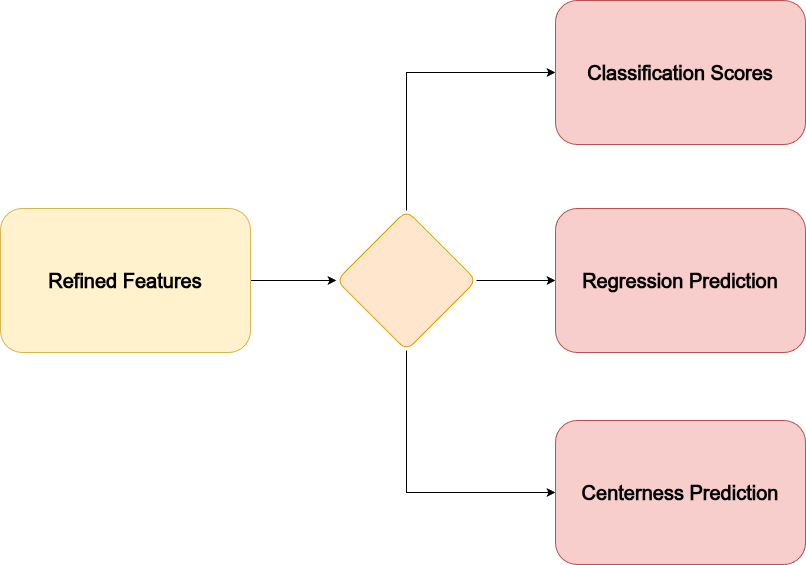}
\caption{Architecture of the Anchor-Free Detection Head. Refined multi-scale features are processed to simultaneously predict classification scores, regression offsets, and centerness values for each pixel.}

\label{fig:classification_head}
\end{figure}

\subsubsection{Loss Function}
Following the formulation proposed in \cite{tian2019fcos}, we utilize a multi-component loss function for our head module to optimize detection performance. This approach combines a classification loss, which employs Focal Loss to handle class imbalance; a centerness loss, which implements Binary Cross-Entropy (BCE) to improve the accuracy of the predicted centerness; and a regression loss, which utilizes Generalized Intersection over Union (GIoU) to increase the precision of the predicted bounding boxes. The total loss is defined as:

\begin{equation}
L_{total} = \lambda_{cls} \cdot L_{focal} + \lambda_{reg} \cdot L_{GIoU} + \lambda_{cent} \cdot L_{BCE}
\end{equation}

In this implementation, we assign a significantly higher weight to the regression component ($\lambda_{reg} = 2.0$) compared to classification and centerness ($\lambda_{cls} = 1.0, \lambda_{cent} = 1.0$). This weighting strategy is motivated by the fact that precise localization is significantly more difficult in high-resolution satellite imagery than object classification. While identifying a "tree" versus "background" is relatively straightforward for the encoder, defining the exact geometric boundaries of small, fragmented, or shadowed tree crowns requires much finer spatial precision.

\section{Experiments}

\subsection{Training Configuration} \label{sec:training_sec}

Our tree detection model was trained for up to 50 epochs using the AdamW optimiser. The initial learning rate was $1 \times 10^{-4}$, and the weight decay was $1 \times 10^{-4}$. We used a cosine-annealing learning rate schedule. Training started with a two-epoch linear warmup phase. The learning rate decayed to a floor of $1 \times 10^{-6}$. Gradient norms were clipped at 1.0 to stabilise training. For data splitting, we retained the source's approach. The training set was divided at the image level: 85\% for training and 15\% for validation. This produced 7,504 training images (63,440 tiles) and 1,319 validation images (11,401 tiles). Input tiles of $640 \times 640$ pixels were used in mini-batches of eight. We applied early stopping with a patience of 10 epochs, monitoring validation mAP. The best checkpoint was saved at epoch 18.
\subsection{Hardware and Software Setup}

All experiments were conducted on a single NVIDIA GeForce RTX 3060 GPU (12 GB), an 8-core CPU, and approximately 64 GB of system memory. The software stack comprised Python 3.11, PyTorch 2.2 with CUDA 12.1, and Albumentations 1.3 for data augmentation.

\subsection{Evaluation Metrics}

Model performance on the held-out validation split was assessed using four complementary metrics. Precision and Recall quantify, respectively, the fraction of predicted bounding boxes that are correct and the fraction of ground-truth trees successfully detected, both evaluated at an intersection-over-union (IoU) threshold of 0.50. Mean Average Precision at IoU 0.50 (mAP@0.50) summarises detection quality by computing the area under the precision–recall curve using COCO-style 101-point interpolation, where precision is interpolated at 101 equally spaced recall levels and averaged. The stricter mAP@0.50-0.95 extends this computation across 10 IoU thresholds from 0.50 to 0.95 in steps of 0.05, reporting their mean, penalising detections that localise tree crowns imprecisely. Together, these metrics, along with the loss, capture both the model's ability to find trees and the spatial accuracy of its predictions (mAP@0.50:0.95).

\section{Results and discussion}

\subsection{Model Performance}
Our model was trained over three days. Training stopped at epoch 20, while the best performance on the validation dataset was achieved at epoch 18.

\begin{figure}[htbp]
\centering
\includegraphics[width=\columnwidth]{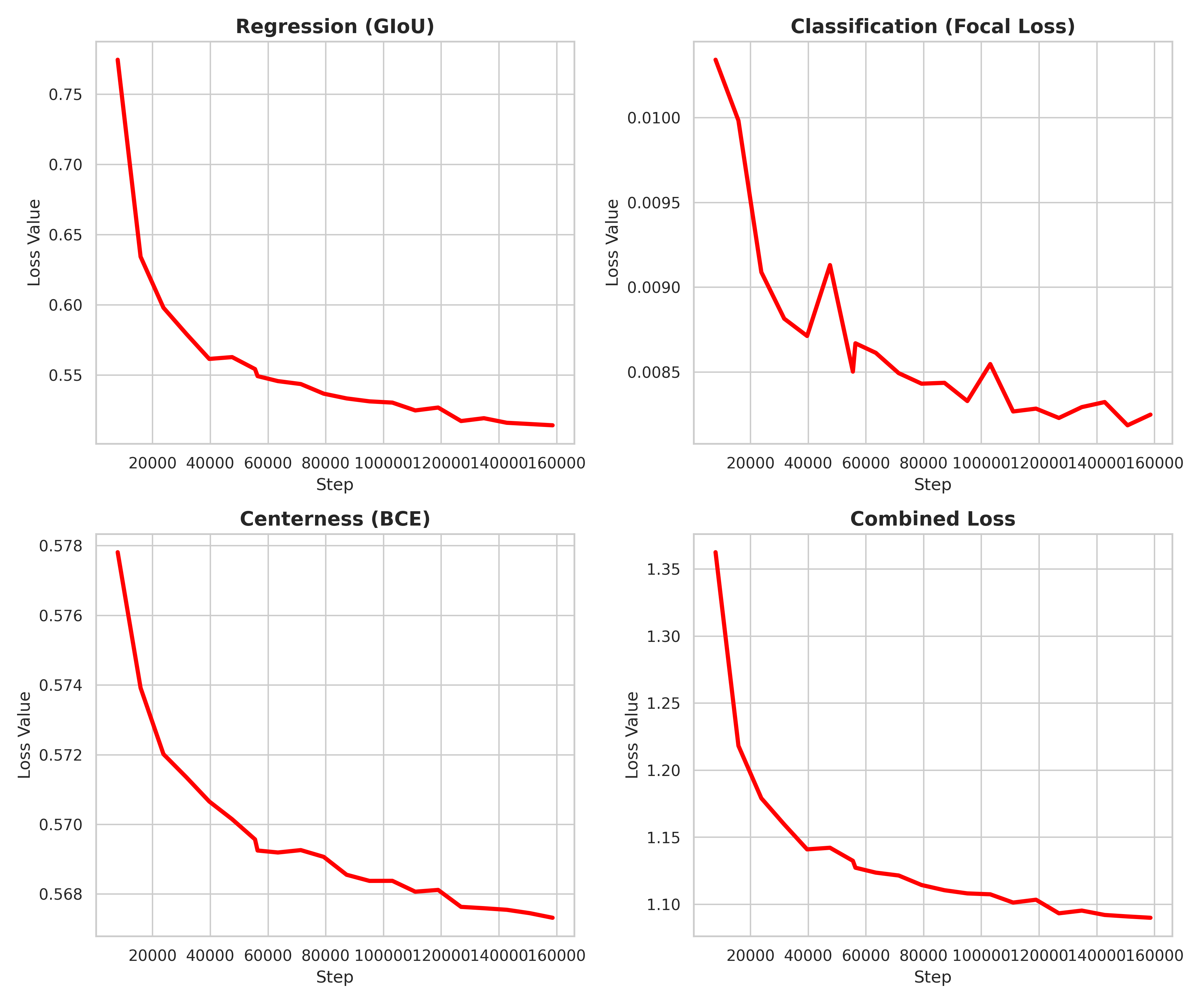}
\caption{Training loss curves for the detection model over time. The charts illustrate the steady convergence of individual components—Regression (GIoU), Classification (Focal Loss), and Centerness (BCE)—as well as the total combined loss.}
\label{fig:loss}
\end{figure}

Fig. \ref{fig:loss} demonstrates stable, continuous convergence of the tree detection model across both the regression and centerness losses, indicating that the model improved its learned tree boundary predictions and object centerness in complex backgrounds over time. We also observe that the focal loss showed some instability, with slight increases and decreases during the early stages of training. However, it began to decline toward the end of training, suggesting that the model initially had difficulty distinguishing tree pixels from background pixels but gradually learned more discriminative patterns to separate trees from other objects. Finally, the combined loss shows a clear trend, with no obvious signs of overfitting.

\begin{figure}[htbp]
  \centering
  \begin{subfigure}[b]{0.30\textwidth}
    \centering
    \includegraphics[width=\columnwidth]{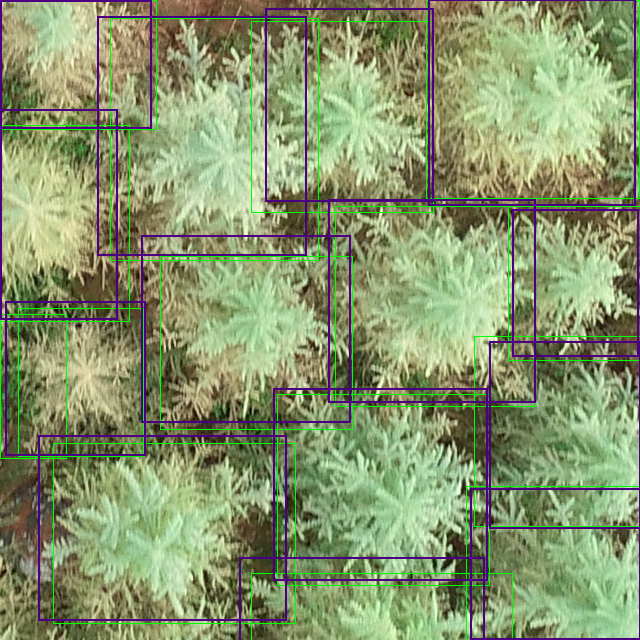}
    \caption{Sample Region A}
  \end{subfigure}
  \hfill
  \begin{subfigure}[b]{0.30\textwidth}
    \centering
    \includegraphics[width=\columnwidth]{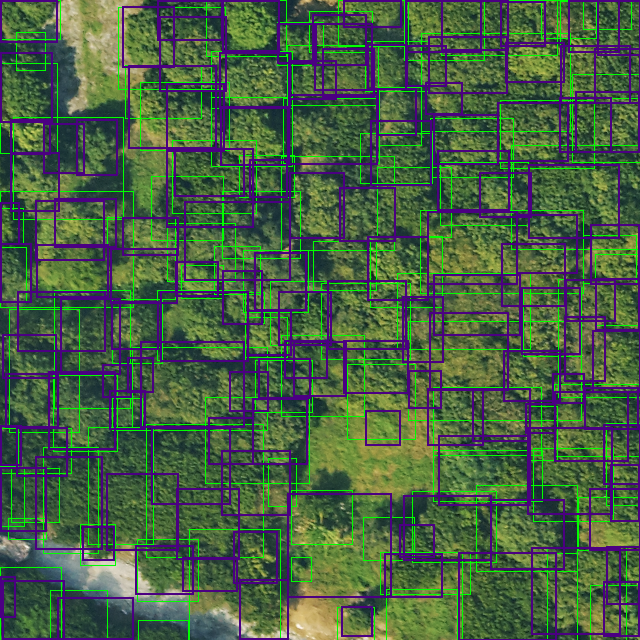}
    \caption{Sample Region B}
  \end{subfigure}
  
  \caption{\textbf{Visual results from the validation dataset of Model Detections. } The images display ground-truth labels (\textbf{green}) and our model's predictions (\textbf{purple}). The high Recall (82\%) is evident as the purple boxes successfully intersect with the vast majority of green boxes.}
  \label{fig:sample_detections}
\end{figure}

Furthermore, our model demonstrates superior performance in identifying tree crowns compared to ground-truth data. As shown in Table \ref{tab:comparison}, our proposed approach achieves a high recall of 82\%, significantly outperforming both the DeepForest baseline (53\%) and the SAM3 model (20.6\%) on the MillionTrees Project benchmark. This sensitivity is visually represented in Figure \ref{fig:sample_detections}; the dense coverage of predicted purple boxes ensures that nearly every green ground-truth box is captured. While this results in a lower precision of 13.89\% due to false positives in complex terrain, the mAP@0.50 of 63.03\% confirms that the model accurately localizes the centers of the actual tree crowns.

\begin{table}[htbp]
\caption{Model Performance Metrics on Validation Set}
\begin{center}
\begin{tabular}{|l|c|}
\hline
\textbf{Metric} & \textbf{Value} \\
\hline
Precision (P) & 13.89\% \\
\hline
Recall (R) & \textbf{82\%} \\
\hline
mAP@0.50 & \textbf{63.03\%} \\
\hline
mAP@0.50:0.95 & 35.47\% \\
\hline
\end{tabular}
\label{tab:metrics}
\end{center}
\end{table}

\begin{table}[htbp]
\caption{Comparative Recall Performance on TreeBoxes Dataset from Milliontrees project}
\begin{center}
\begin{tabular}{|l|c|}
\hline
\textbf{Model} & \textbf{Avg. Recall} \\
\hline
SAM3 & 20.6\% \\
\hline
DeepForest baseline & 53\% \\
\hline
\textbf{Ours (Proposed)} & \textbf{82\%} \\
\hline
\end{tabular}
\label{tab:comparison}
\end{center}
\end{table}

\section{Practical Implementation in the Palestinian Context}
We believe this AI model will serve as the foundation for creating a custom satellite imagery dataset for Palestine, thereby facilitating the development of various AI tasks beyond simple tree counting and detection. Our goal is to expand these capabilities to include tree species identification, such as olive, almond, and palm, as well as the measurement of tree height and canopy width.

To achieve this, we developed a web application that serves as an annotation tool, supported by an AI model, to verify every tree in Palestine and assign a unique ID. The system generates reports on the expansion or decline of specific tree types and overall green cover within targeted areas. We built the application on the Esri mapping framework because it effectively serves our initial proof of concept, even though it may not provide the highest possible map resolution. Additionally, we integrated the application with GeoMolg, developed by the Palestinian Ministry of Local Government. This allows us to leverage vital geospatial data such as governorate land boundaries, community boundaries, and registered blocks and parcels, all integrated as polygons within our system.

Our work functions at three distinct levels: Scene Detection, Parcel Detection, and Community Detection.

\subsection{Scene Detection}
\begin{figure}[htbp]
\centering
\includegraphics[width=\columnwidth]{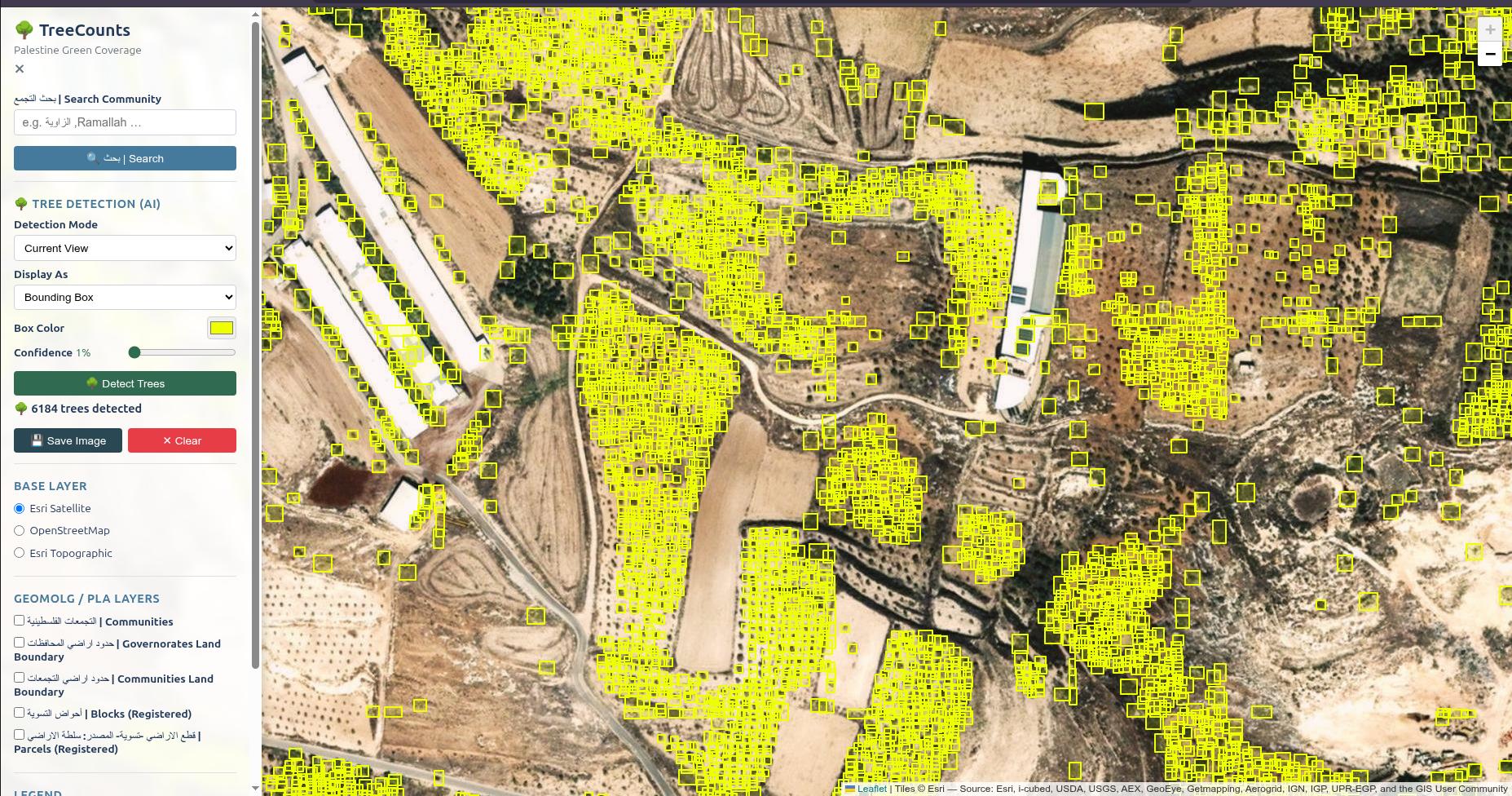}
\caption{Visual representation of the 6,184 trees detected by the model, where each yellow point represents a bounding box for a tree identified with a confidence threshold of 1\%}
\label{fig:Scene}
\end{figure}

The first level analyzes the current map viewport displayed on the screen, enabling rapid assessment of any visible area, regardless of administrative boundaries. This approach is ideal for quick surveys or exploratory analysis of arbitrary regions as shown in Fig \ref{fig:Scene}. In this instance, the model detected 6,184 trees in the current scene, a significant number of detections. While the model did not identify some trees, one of its most impressive features is that it recognized cultivated cropland and correctly excluded it from the tree count.

\subsection{Parcel Detection}
\begin{figure}[htbp]
\centering
\includegraphics[width=\columnwidth]{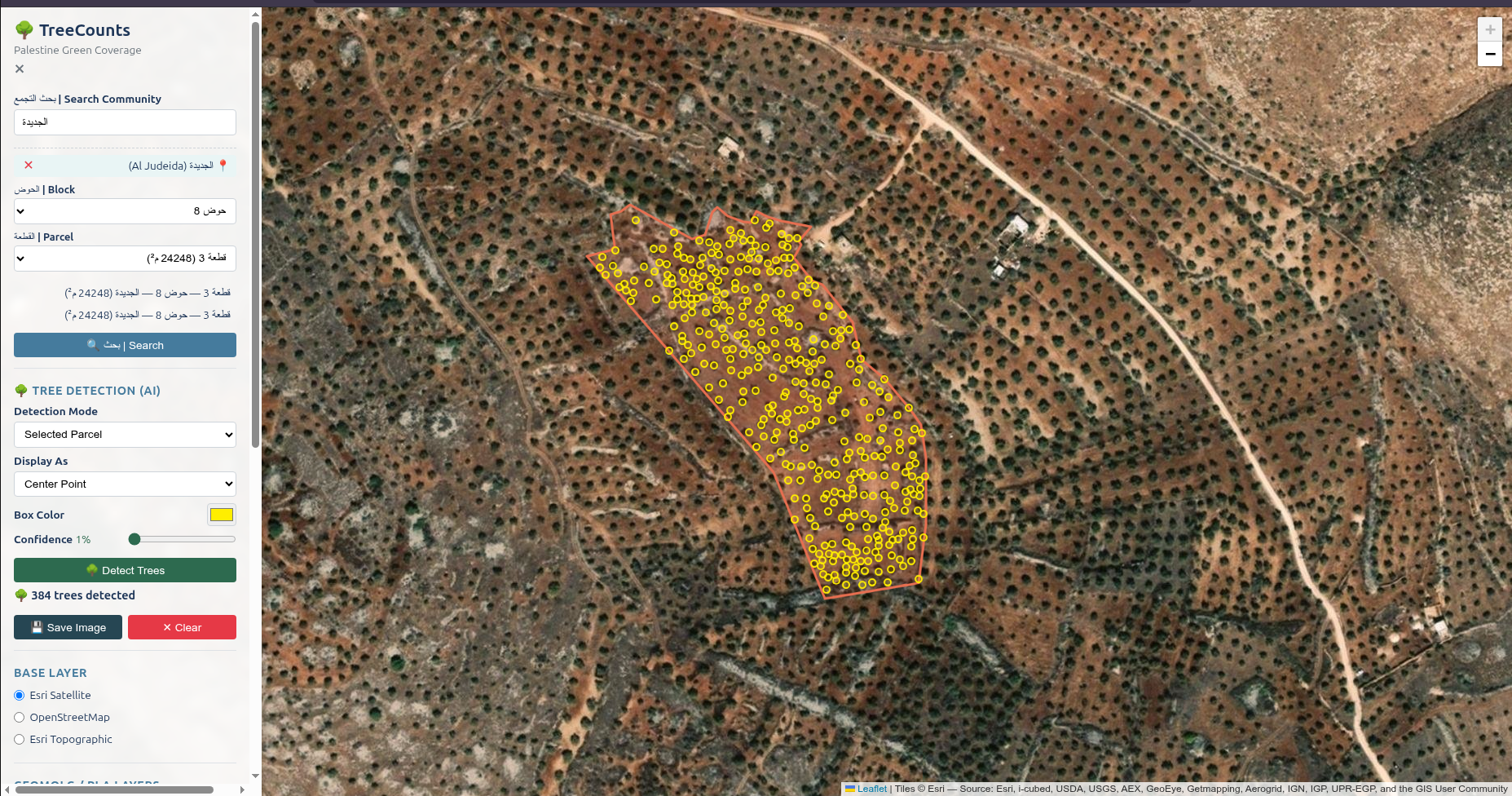}
\caption{Visual representation of the 384 trees detected by the model on a parcel containing approximately 410 trees total. The model maintained accuracy despite the presence of stone chains, where each yellow point represents a bounding box identified with a confidence threshold of 1\%}
\label{fig:parcel}
\end{figure}

The second level operates at the level of individual registered parcels. Each parcel in Palestine is registered with a specific ID and defined boundaries, which we retrieved from the GeoMolg system. When a user selects a specific community, block, and parcel through the search feature, the model processes the satellite imagery within the exact polygon boundaries of that parcel and clips all detections to the outline. This produces a precise tree count tied to a legally defined land unit, which is essential for agricultural reporting and land use documentation. In Fig. \ref{fig:parcel}, we tested the model on a parcel known to contain approximately 400 olive trees and 10 almond trees. Despite the presence of multiple stone chains and the fact that this area consists of unseen data with limited image quality, the model successfully detected 384 trees. This represents a promising result for the system.

\subsection{Community Detection}
\begin{figure}[htbp]
\centering
\includegraphics[width=\columnwidth]{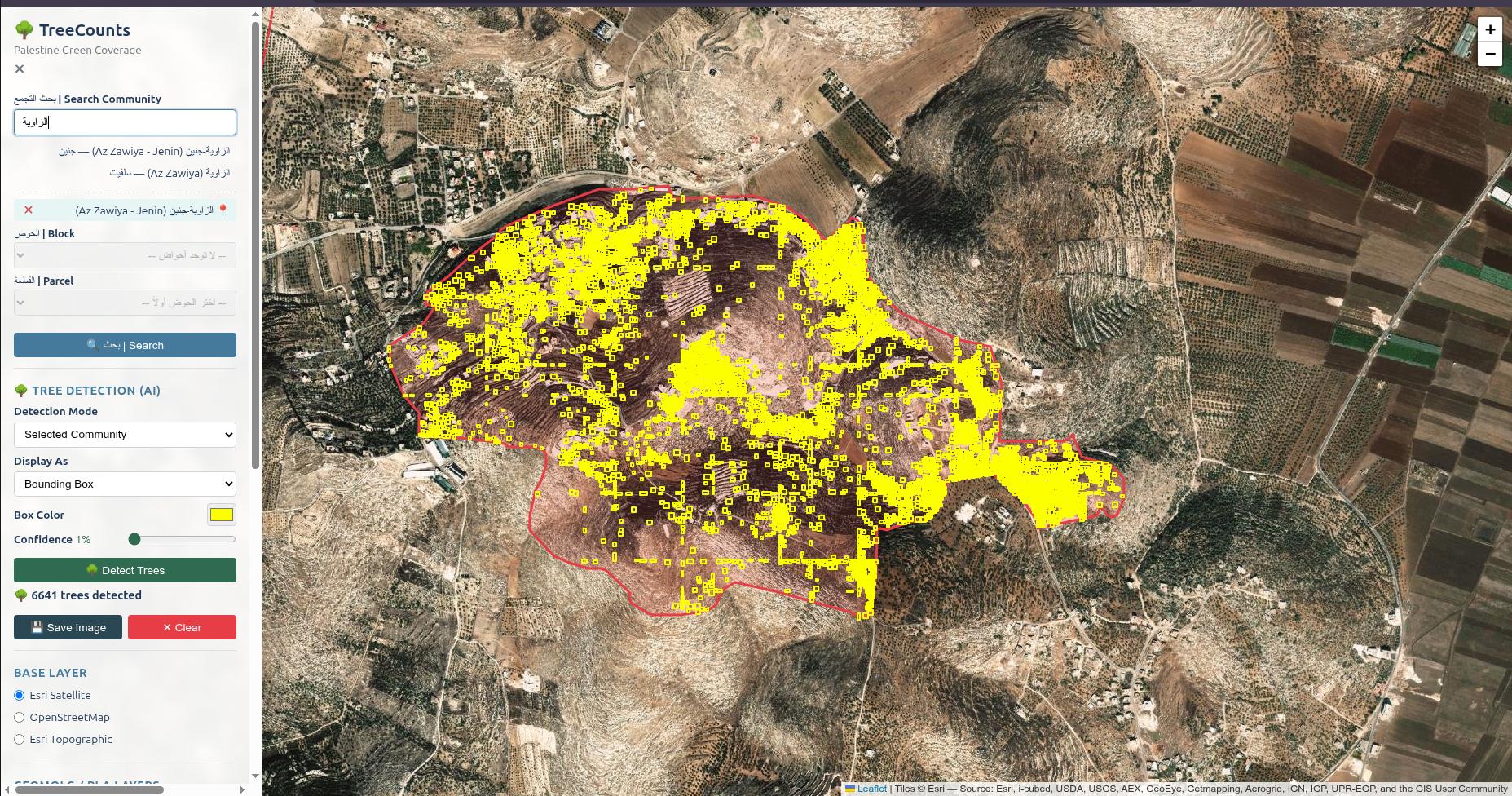}
\caption{Visual representation of the 6,641 trees detected by the model in Al Zawiya using a confidence threshold of 1\%}
\label{fig:community}
\end{figure}

The third and most comprehensive level processes the land boundary of an entire community. Because communities can span large geographic areas containing thousands of satellite tiles, the system divides the polygon into spatial chunks, processes each chunk independently through the inference pipeline, and then merges the results across chunk boundaries. Progress is streamed back to the user in real time, showing which chunk is being processed out of the total. This hierarchical approach, from a single screen view to a single parcel and finally to an entire community, allows stakeholders to operate at the scale most relevant to their needs. We tested this method on one of the smallest villages in the West Bank called Al Zawiya. As shown in Fig. \ref{fig:community}, the model detected 6,641 trees across the community. While this number may not be perfectly accurate due to some trees being missed because of low image quality, it provides essential high-level indicators for each community.

\section{Conclusion}
In this study, we developed a specialized deep learning pipeline for tree detection in satellite imagery, specifically designed to address the challenges posed by fragmented agricultural landscapes and resource-constrained environments such as Palestine. By integrating a ResNet-50 backbone with Atrous Spatial Pyramid Pooling (ASPP) and an anchor-free FCOS detection head, our model achieved a significant recall of 82\% on the MillionTrees benchmark, substantially outperforming existing baselines. Furthermore, we successfully transitioned from theoretical modeling to practical implementation by developing a web-based geographic information system (GIS) integrated with Palestinian cadastral data. Our results demonstrate that satellite-based AI tools can provide a scalable, cost-effective alternative to traditional agricultural census methods, offering critical data-driven insights for policy-makers and farmers in regions where aerial surveillance is restricted.

\section{Future Work}
Building upon the current findings, several avenues for future research and technical enhancement will be pursued:


\begin{itemize}
    \item We would like to experiment with applying a dynamic threshold to our model. In satellite imagery, image quality is often low and tree density can vary significantly across images, so the optimal threshold may also vary. Therefore, as part of future work to improve this pipeline, we plan to adopt a dynamic thresholding approach.

    \item In this study, the model was trained with an input size of $640 \times 640$ pixels due to hardware limitations. Therefore, as part of future work , we plan to enlarge the architecture on high-performance computing infrastructure with larger GPU capacity, which would enable the use of larger input sizes and may further improve detection performance.

    \item The current pipeline focuses only on tree detection. However, in the Palestinian context, distinguishing between tree species such as olive, almond, and palm trees would provide additional practical value for agricultural monitoring and planning. Therefore, as part of future work to improve this pipeline, we plan to extend the framework by incorporating a classification head to identify different tree species.

    \item In future work, we plan to use our developed web application to build a custom human verified dataset from the Palestinian landscape. Such a dataset would better capture local topography, agricultural layouts, and tree distribution patterns, providing a more suitable basis for fine-tuning the model.
    
\end{itemize}



\bibliographystyle{IEEEtran}
\bibliography{IEEEexample}

\end{document}